\documentclass[10pt,twocolumn,letterpaper]{article}

\usepackage{cvpr}
\usepackage{times}
\usepackage{epsfig}
\usepackage{graphicx}
\usepackage{amsmath}
\usepackage{amssymb}
\usepackage[usenames,dvipsnames]{xcolor}

\usepackage[pagebackref=true,breaklinks=true,letterpaper=true,colorlinks,bookmarks=false]{hyperref}

\cvprfinalcopy 

\def\cvprPaperID{****} 

\ifcvprfinal\fi
\begin{document}

\onecolumn

\vspace*{20 mm}

\begin{center}
{\huge UCF101: A Dataset of 101 Human Actions }
\end{center}

\begin{center}
{\huge Classes From Videos in The Wild}
\end{center}

\begin{center}
Khurram Soomro, Amir Roshan Zamir and Mubarak Shah
\end{center}

\begin{center}
CRCV-TR-12-01
\end{center}

\begin{center}
November 2012
\end{center}

\vspace{120 mm}

\begin{center}
\end{center}

\vspace{5 mm}

\begin{center}
Center for Research in Computer Vision
\end{center}

\begin{center}
University of Central Florida
\end{center}

\begin{center}
4000 Central Florida Blvd.
\end{center}

\begin{center}
Orlando, FL 32816-2365 USA
\end{center}

\twocolumn

\title{UCF101: A Dataset of 101 Human Actions Classes From Videos in The Wild}
\author{Khurram Soomro, Amir Roshan Zamir and Mubarak Shah\\
Center for Research in Computer Vision, Orlando, FL 32816, USA\\
{\tt\ $\{$ksoomro, aroshan, shah$\}$@cs.ucf.edu}\\
{\url{http://crcv.ucf.edu/data/UCF101.php}}
}

\maketitle

\begin{abstract}
We introduce UCF101 which is currently the largest dataset of human actions.
It consists of 101 action classes, over 13k clips and 27 hours of video data.
The database consists of realistic user-uploaded videos containing camera motion and cluttered background.
Additionally, we provide baseline action recognition results on this new dataset using standard bag of words approach with overall performance of 44.5\%.
To the best of our knowledge, UCF101 is currently the most challenging dataset of actions due to its large number of classes, large number of clips and also unconstrained nature of such clips.
\end{abstract}


\section{Introduction}

The majority of existing action recognition datasets suffer from two disadvantages: \textbf{1)} The number of their classes is typically very low compared to the richness of performed actions by humans in reality, e.g. KTH \cite{KTH}, Weizmann \cite{Weizmann}, UCF Sports \cite{UCFSports}, IXMAS \cite{IXMAS} datasets includes only 6, 9, 9, 11 classes respectively. \textbf{2)} The videos are recorded in unrealistically controlled environments. For instance, KTH, Weizmann, IXMAS are staged by actors; HOHA \cite{HOHA} and UCF Sports are composed of movie clips captured by professional filming crew. Recently, web videos have been used in order to utilize unconstrained user-uploaded data to alleviate the second issue \cite{UCF11,Olympic,UCF50,HMDB51}. However, the first disadvantage remains unresolved as the largest existing dataset does not include more than 51 actions while several works showed that the number of classes play a crucial role in evaluating an action recognition method \cite{Johansson,UCF50}. Therefore, we have compiled a new dataset with 101 actions and 13320 clips which is nearly twice bigger than the largest existing dataset in terms of number of actions and clips. (HMDB51 \cite{HMDB51} and UCF50 \cite{UCF50} are the currently the largest ones with 6766 clips of 51 actions and 6681 clips of 50 actions respectively.)

\begin{figure}
\begin{center}
   \includegraphics[width=1\linewidth]{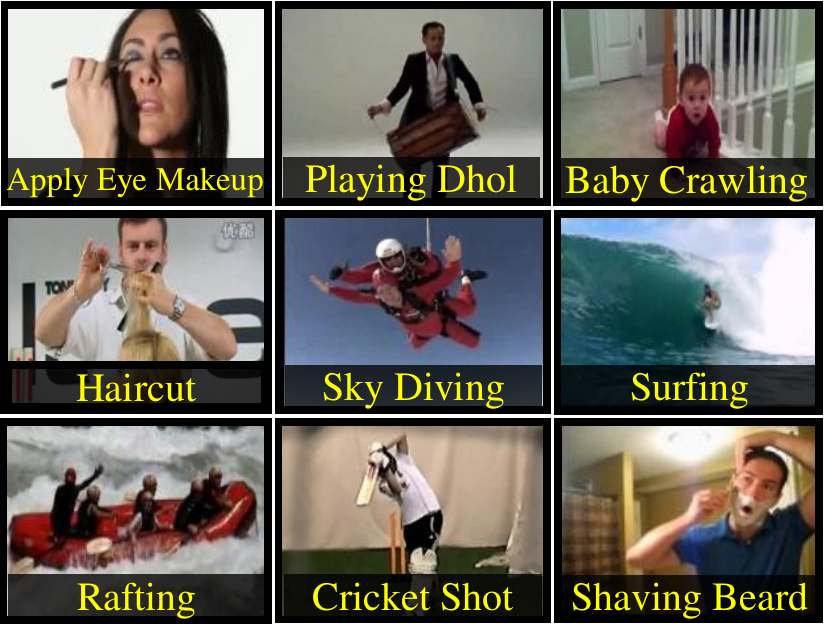}
\end{center}
   \caption{Sample frames for 6 action classes of UCF101.}
\label{fig:firstpage}
\end{figure}

The dataset is composed of web videos which are recorded in unconstrained environments and typically include
camera motion, various lighting conditions, partial occlusion, low quality frames, etc. Fig. \ref{fig:firstpage} shows sample frames of 6 action classes from UCF101.

\section{Dataset Details}

\begin{figure*}
\begin{center}
   \includegraphics[width=1\linewidth]{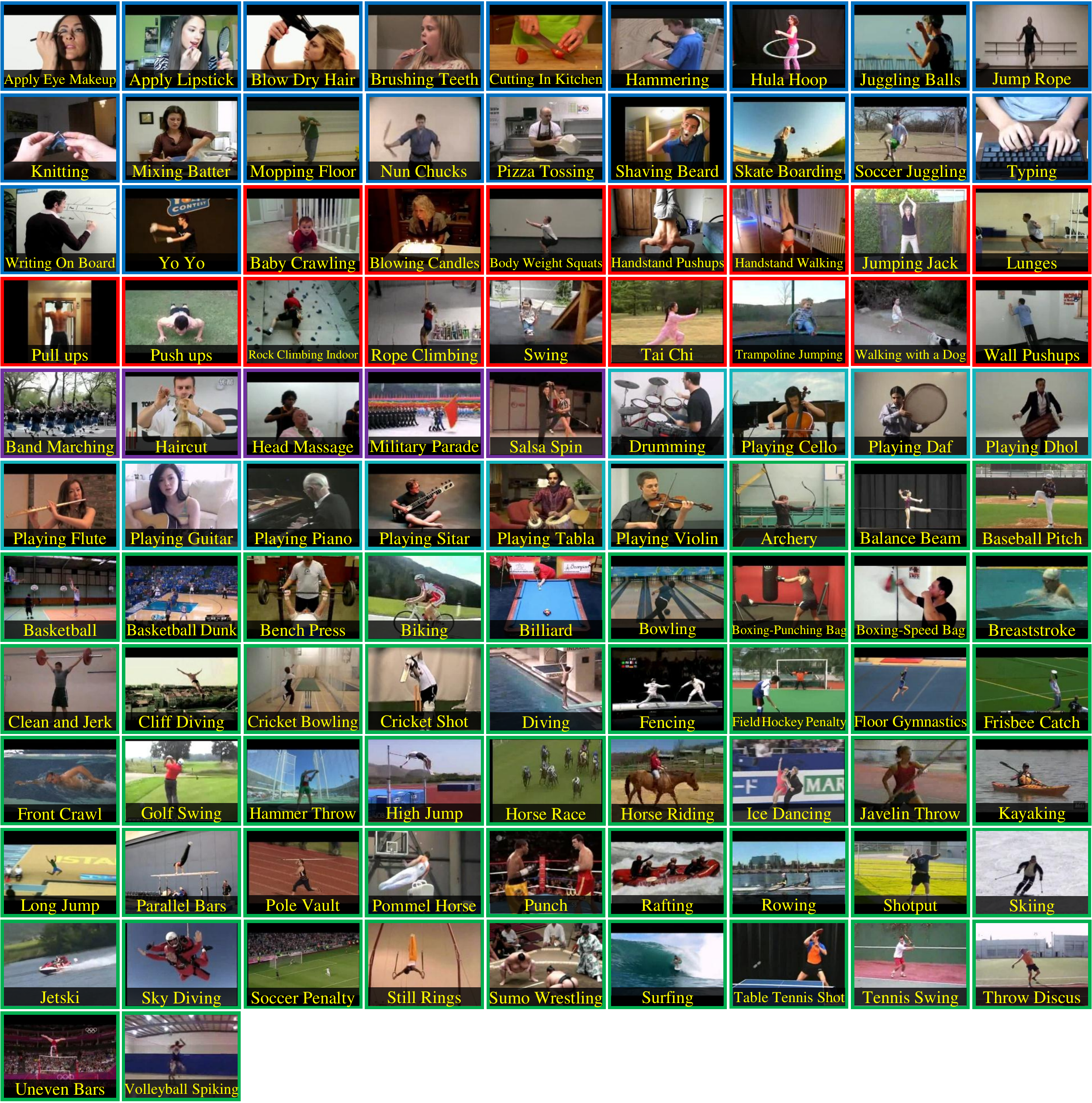}
\end{center}
   \caption{101 actions included in UCF101 shown with one sample frame. The color of frame borders specifies to which action type they belong: {\color{NavyBlue} Human-Object Interaction}, {\color{red} Body-Motion Only}, {\color{RoyalPurple} Human-Human Interaction}, {\color{YellowOrange} Playing Musical Instruments}, {\color{ForestGreen} Sports}.}
\label{fig:frames}
\end{figure*}

\textbf{Action Classes:} UCF101 includes total number of 101 action classes which we have divided into five types: {\color{NavyBlue}Human-Object Interaction}, {\color{red}Body-Motion Only}, {\color{RoyalPurple} Human-Human Interaction}, {\color{YellowOrange}Playing Musical Instruments}, {\color{ForestGreen}Sports}.

UCF101 is an extension of UCF50 which included the following 50 action classes:
\emph{\{{\color{ForestGreen}Baseball Pitch, Basketball Shooting, Bench Press, Biking, Billiards Shot, Breaststroke, Clean and Jerk, Diving,} {\color{YellowOrange}Drumming}, {\color{ForestGreen}Fencing, Golf Swing, High Jump, Horse Race, Horse Riding,} {\color{NavyBlue}Hula Hoop}, {\color{ForestGreen}Javelin Throw,}, {\color{NavyBlue}Juggling Balls}, {\color{red}Jumping Jack}, {\color{NavyBlue}Jump Rope}, {\color{ForestGreen}Kayaking,} {\color{red}Lunges}, {\color{RoyalPurple}Military Parade}, {\color{NavyBlue}Mixing Batter}, {\color{NavyBlue}Nun chucks}, {\color{NavyBlue}Pizza Tossing}, {\color{YellowOrange}Playing Guitar}, {\color{YellowOrange}Playing Piano}, {\color{YellowOrange}Playing Tabla}, {\color{YellowOrange}Playing Violin}, {\color{ForestGreen}Pole Vault, Pommel Horse,} {\color{red}Pull Ups}, {\color{ForestGreen}Punch,} {\color{red}Push Ups}, {\color{red}Rock Climbing Indoor}, {\color{red}Rope Climbing}, {\color{ForestGreen}Rowing,} {\color{RoyalPurple}Salsa Spins}, {\color{NavyBlue}Skate Boarding}, {\color{ForestGreen}Skiing, Skijet,} {\color{NavyBlue}Soccer Juggling}, {\color{red}Swing}, {\color{red}TaiChi}, {\color{ForestGreen}Tennis Swing}, {\color{ForestGreen}Throw Discus}, {\color{red}Trampoline Jumping}, {\color{ForestGreen}Volleyball Spiking,} {\color{red}Walking with a dog}, {\color{NavyBlue}Yo Yo}\}.}
The color class labels specify\ which predefined action type they belong to.

\begin{figure*}
\begin{center}
   \includegraphics[width=1\linewidth]{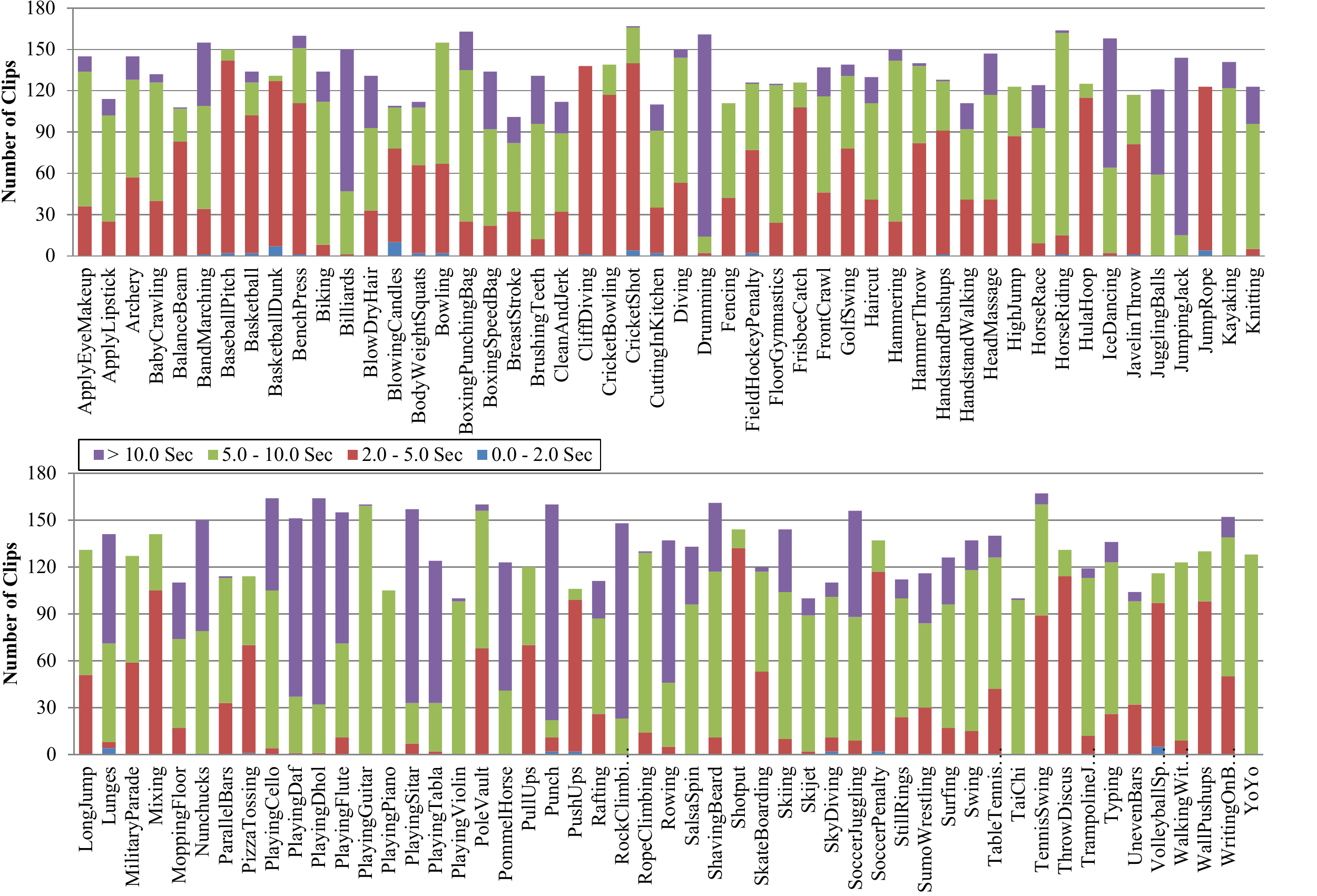}
\end{center}
   \caption{Number of clips per action class. The distribution of clip durations is illustrated by the colors.}
\label{fig:clipchart}
\end{figure*}

\begin{figure*}
\begin{center}
   \includegraphics[width=1\linewidth]{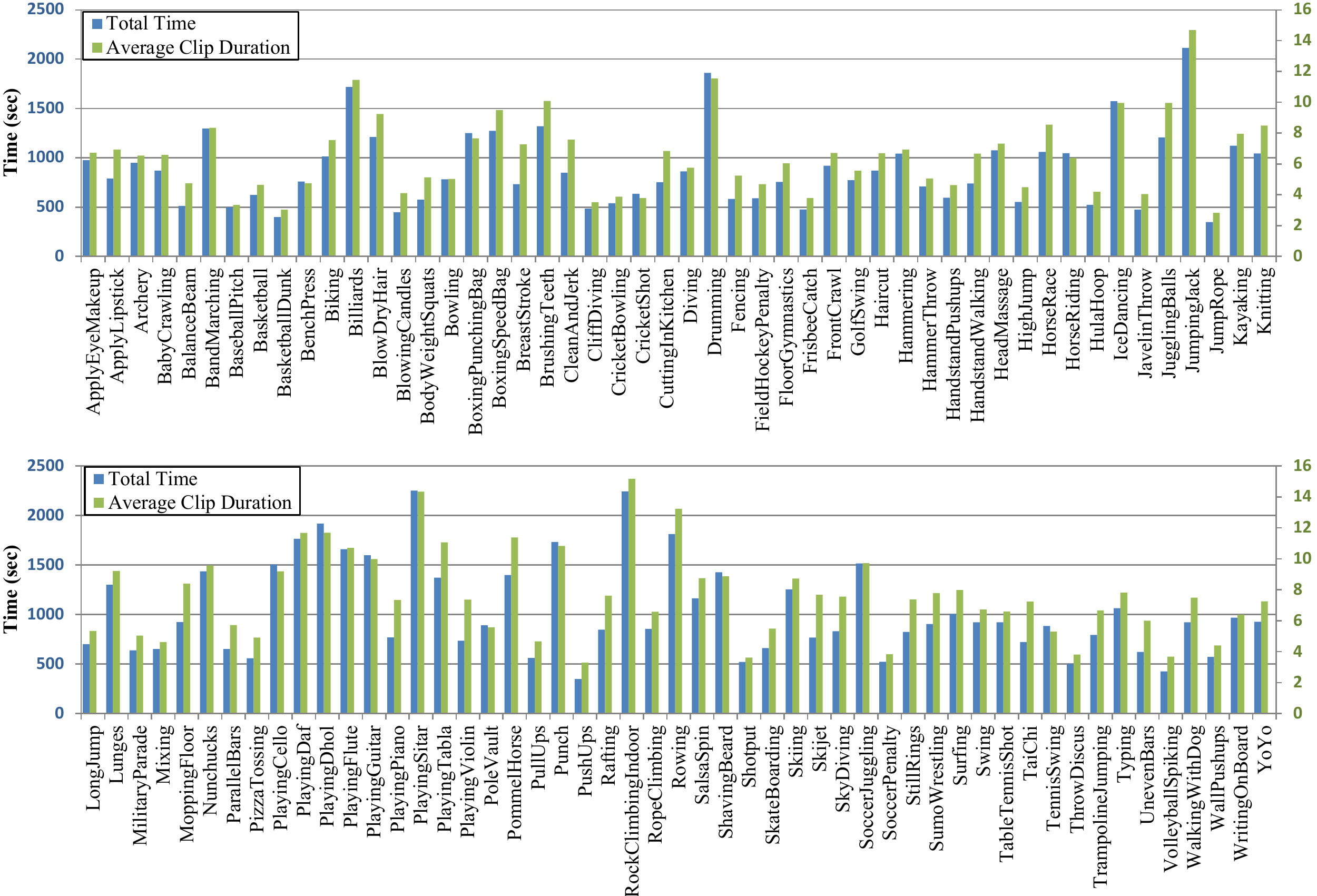}
\end{center}
   \caption{Total time of videos for each class is illustrated using the blue bars. The average length of the clips for each action is depicted in green.}
\label{fig:timechart}
\end{figure*}

The following 51 new classes are introduced in UCF101:
\emph{\{{\color{NavyBlue}Apply Eye Makeup}, {\color{NavyBlue}Apply Lipstick}, {\color{ForestGreen}Archery},  {\color{red}Baby Crawling},  {\color{ForestGreen}Balance Beam},  {\color{RoyalPurple}Band Marching}, {\color{ForestGreen}Basketball Dunk},
{\color{NavyBlue}Blow Drying Hair}, {\color{red}Blowing Candles}, {\color{red}Body Weight Squats}, {\color{ForestGreen}Bowling},{\color{ForestGreen}Boxing-Punching Bag}, {\color{ForestGreen}Boxing-Speed Bag}, {\color{NavyBlue}Brushing Teeth},  {\color{ForestGreen}Cliff Diving}, {\color{ForestGreen}Cricket Bowling}, {\color{ForestGreen}Cricket Shot}, {\color{NavyBlue}Cutting In Kitchen}, {\color{ForestGreen}Field Hockey Penalty}, {\color{ForestGreen}Floor Gymnastics}, {\color{ForestGreen}Frisbee Catch}, {\color{ForestGreen}Front Crawl}, {\color{RoyalPurple}Hair cut}, {\color{NavyBlue}Hammering}, {\color{ForestGreen}Hammer Throw}, {\color{red}Handstand Pushups}, {\color{red}Handstand Walking}, {\color{RoyalPurple}Head Massage}, {\color{ForestGreen}Ice Dancing}, {\color{NavyBlue}Knitting}, {\color{ForestGreen}Long Jump}, {\color{NavyBlue}Mopping Floor}, {\color{ForestGreen}Parallel Bars}, {\color{YellowOrange}Playing Cello}, {\color{YellowOrange}Playing Daf}, {\color{YellowOrange}Playing Dhol}, {\color{YellowOrange}Playing Flute}, {\color{YellowOrange}Playing Sitar}, {\color{ForestGreen}Rafting}, {\color{NavyBlue}Shaving Beard}, {\color{ForestGreen}Shot put}, {\color{ForestGreen}Sky Diving}, {\color{ForestGreen}Soccer Penalty}, {\color{ForestGreen}Still Rings}, {\color{ForestGreen}Sumo Wrestling}, {\color{ForestGreen}Surfing}, {\color{ForestGreen}Table Tennis Shot}, {\color{NavyBlue}Typing}, {\color{ForestGreen}Uneven Bars}, {\color{red}Wall Pushups}, {\color{NavyBlue}Writing On Board}\}.
}
Fig. \ref{fig:frames} shows a sample frame for each action class of UCF101.

\textbf{Clip Groups: }The clips of one action class are divided into 25 groups which contain 4-7 clips each. The clips in one group share some common features, such as the background or actors.

The bar chart of Fig. \ref{fig:clipchart} shows the number of clips in each class. The colors on each bar illustrate the durations of different clips included in that class. The chart shown in Fig. \ref{fig:timechart} illustrates the average clip length (green) and total duration of clips (blue) for each action class.

The videos are downloaded from YouTube \cite{YouTube} and the irrelevant ones are manually removed. All clips have fixed frame rate and resolution of 25 FPS and $320\times240$ respectively. The videos are saved in {\tt .avi} files compressed using \emph{DivX} codec available in k-lite package \cite{klite}. The audio is preserved for the clips of the new 51 actions. Table \ref{tab:summary} summarizes the characteristics of the dataset.

\begin{table}
\begin{center}
\begin{tabular}{|c||c|}
\hline
Actions & 101 \\
\hline
Clips  & 13320\\
\hline
Groups per Action & 25  \\
\hline
Clips per Group  & 4-7\\
\hline
Mean Clip Length & 7.21 sec  \\
\hline
Total Duration   & 1600 mins \\
\hline
Min Clip Length & 1.06 sec  \\
\hline
 Max Clip Length  & 71.04 sec\\
\hline
Frame Rate & 25 fps  \\
\hline
 Resolution  & 320$\times$240\\
\hline
 Audio  & Yes (51 actions)\\
\hline
\end{tabular}
\end{center}
\caption{Summary of Characteristics of UCF101}
\label{tab:summary}
\end{table}

\textbf{Naming Convention:} The zipped file of the dataset (available at \url{http://crcv.ucf.edu/data/UCF101.php} ) includes 101 folders each containing the clips of one action class. The name of each clip has the following form:

\begin{equation}
{\tt v\_\textbf{X}\_g\textbf{Y}\_c\textbf{Z}.avi }
\nonumber
\end{equation}
where \textbf{X}, \textbf{Y} and \textbf{Z} represent action class label, group and clip number respectively. For instance, {\tt v\_ApplyEyeMakeup\_g03\_c04.avi } corresponds to the clip 4 of group 3 of action class {\tt ApplyEyeMakeup}.

\begin{table*}
\begin{center}
\begin{tabular}{ c|c c c c c c}
\hline
Dataset & Number of Actions & Clips & Background & Camera Motion & Release Year & Resource \\
\hline
\hline

KTH \cite{KTH}   & 6 & 600 & Static & Slight & 2004 & Actor Staged\\
Weizmann \cite{Weizmann}  & 9 &  81 & Static & No & 2005 & Actor Staged\\
UCF Sports \cite{UCFSports}  & 9 &  182 & Dynamic & Yes & 2009 & TV, Movies\\
IXMAS \cite{IXMAS} & 11 &  165 & Static & No & 2006 & Actor Staged\\
UCF11 \cite{UCF11}  & 11 & 1168 & Dynamic & Yes & 2009 & YouTube\\
HOHA \cite{HOHA}  & 12 &  2517 & Dynamic & Yes & 2009 & Movies\\
Olympic \cite{Olympic}  & 16 &  800 & Dynamic & Yes & 2010 & YouTube\\
UCF50 \cite{UCF50} & 50 &  6681 & Dynamic & Yes & 2010 & YouTube\\
HMDB51 \cite{HMDB51} & 51 & 6766 &Dynamic & Yes & 2011 & Movies, YouTube, Web\\
\hline

\textbf{UCF101}  & \textbf{101} & \textbf{13320} & Dynamic & Yes & \textbf{2012} &  YouTube  \\

\hline
\end{tabular}
\end{center}
\caption{Summary of Major Action Recognition Datasets}
\label{tab:datasets}
\end{table*}

\section{Experimental Results}

We performed an experiment using bag of words approach which is widely accepted as a standard action recognition method to provide baseline results on UCF101.

\begin{figure*}
\begin{center}
   \includegraphics[width=1\linewidth]{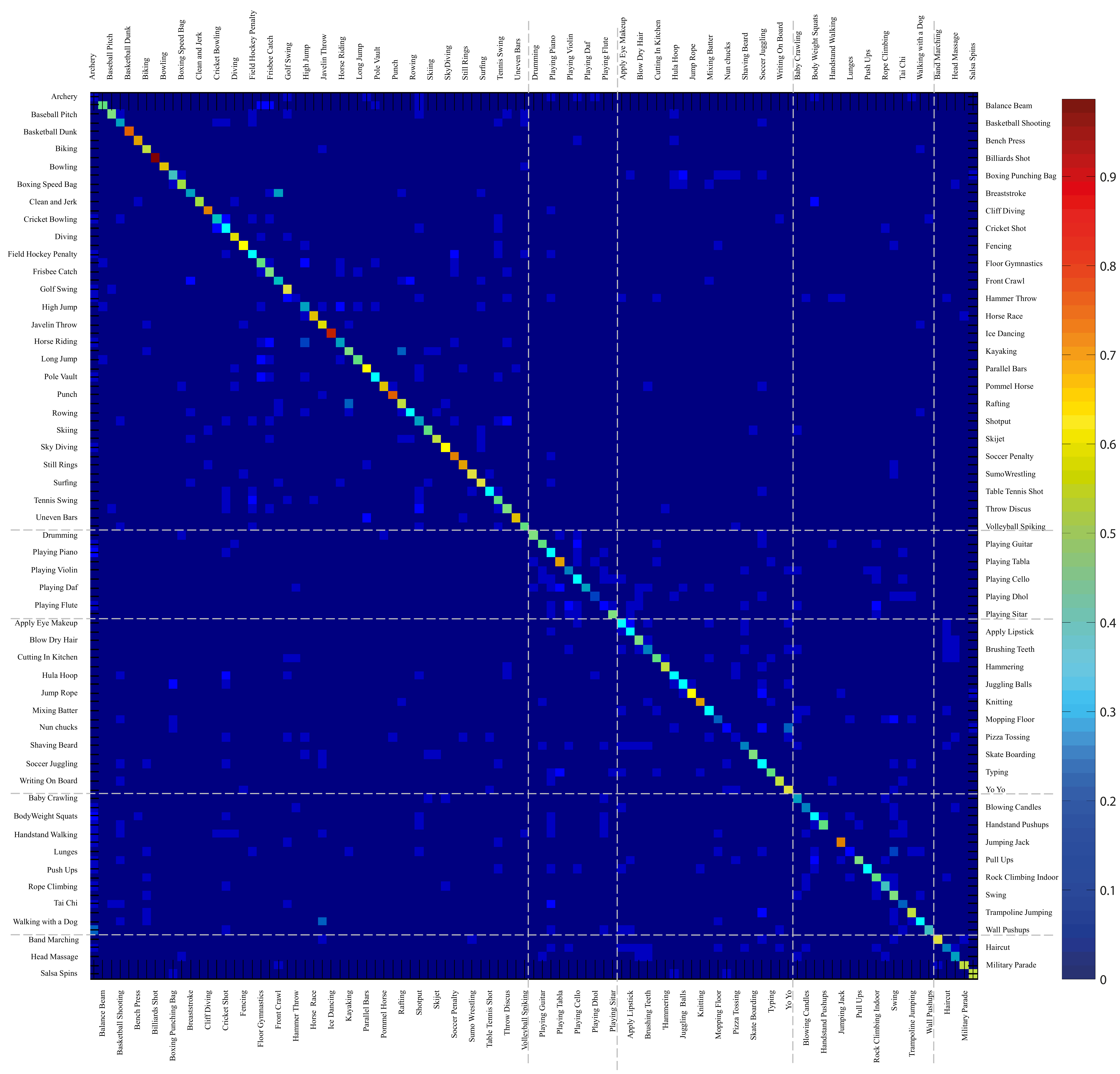}
\end{center}
   \caption{Confusion table of baseline action recognition results using bag of words approach on UCF101. The drawn lines separate different types of actions; 1-50: Sports, 51-60: Playing Musical Instrument, 61-80: Human-Object Interaction, 81-96: Body-Motion Only, 97-101: Human-Human Interaction.}
\label{fig:confusion}
\end{figure*}

From each clip, we extracted Harris3D corners (using the implementation by \cite{HOHA}) and computed 162 dimensional HOG/HOF descriptors for each. We clustered a randomly selected set of 100,000 space-time interest points (STIP) using k-means to build the codebook. The size of our codebook is k=4000 which is shown to yield good results over a wide range of datasets. The descriptors were assigned to their closest video words using nearest neighbor classifier, and each clip was represented by a 4000-dimensional histogram of its words. Utilizing a leave-one-group-out 25-fold cross validation scenario, a SVM was trained using the histogram vectors of the training folds. We employed a nonlinear multiclass SVM with histogram intersection kernel and 101 classes each representing one action. For testing, a similar histogram representation for the query video was computed and classified using the trained SVM. This method yielded an overall accuracy of 44.5\%; The confusion matrix for all 101 actions is shown in Fig. \ref{fig:confusion}.

The accuracy for the predefined action types are:  Sports (50.54\%), Playing Musical Instrument  (37.42\%), Human-Object Interaction (38.52\%), Body-Motion Only (36.26\%), Human-Human Interaction (44.14\%). Sports actions achieve the highest accuracy since performing sports typically requires distinctive motions which makes the classification easier. Moreover, the background in sports clips are generally less cluttered compared to other action types. Unlike Sports Actions, Human-Object Interaction clips typically have a highly cluttered background. Additionally, the informative motions typically occupy a small portion of the motions in the clips which explains the low recognition accuracy of this action class.

We recommend a \emph{25-fold cross validation} experimental setup using all the videos in the dataset to keep consistency of the reported tests on UCF101; the baseline results provided in this section were computed using the same scenario.

\section{Related Datasets}

UCF Sports, UCF11, UCF50 and UCF101 are the four action  datasets compiled by UCF in chronological order; each one includes its precursor. We made two minor modifications in the portion of UCF101 which includes UCF50 videos: the number of groups is fixed to 25 for all the actions, and each group includes up to 7 clips. Table \ref{tab:datasets}  shows a list of existing action recognition datasets with detailed characteristics of each. Note that UCF101 is remarkably larger than the rest.

\section{Conclusion}

We introduced UCF101 which is the most challenging dataset for action recognition compared to the existing ones. It includes 101 action classes and over 13k clips which makes it outstandingly larger than other datasets. UCF101 is composed of unconstrained videos downloaded from YouTube which feature challenges such as poor lighting, cluttered background and severe camera motion. We provided baseline action recognition results on this new dataset using standard bag of words method with overall accuracy of 44.5\%.

{\small
\bibliographystyle{ieee}
\bibliography{egbib}
}

\end{document}